\documentclass{article}

\usepackage[accepted]{icml2021}

\DeclareUnicodeCharacter{2212}{-}

\usepackage{hyperref}       
\usepackage{url}            
\usepackage{booktabs}       
\usepackage{nicefrac}       
\usepackage{microtype}      
\usepackage{amsmath}
\usepackage{graphicx}
\usepackage{subfigure}
\usepackage{tikz}
\usepackage{pgf}
\usepackage{neuralnetwork}
\usetikzlibrary{positioning, fit, arrows.meta, shapes}

\newcommand{\empt}[2]{$#1^{\langle #2 \rangle}$}

\title{Fast-Slow Streamflow Model Using Mass-Conserving LSTM}

\icmltitlerunning{Fast-Slow Streamflow Model Using Mass-Conserving LSTM}
\begin{document}
\twocolumn[
\icmltitle{Fast-Slow Streamflow Model Using Mass-Conserving LSTM}


\icmlsetsymbol{equal}{*}

\begin{icmlauthorlist}
\icmlauthor{Miguel Paredes Qui\~{n}ones}{equal,ibm}
\icmlauthor{Maciel Zortea}{equal,ibm}
\icmlauthor{Leonardo S. A. Martins}{equal,ibm}
\end{icmlauthorlist}

\icmlaffiliation{ibm}{IBM Research, S\~{a}o Paulo, Brasil}

\icmlcorrespondingauthor{Miguel Paredes Qui\~{n}ones}{mparedes@br.ibm.com}

\icmlkeywords{Streamflow forecasting, simulation, LSTM, ICML, Parana river}

\vskip 0.3in
]



\printAffiliationsAndNotice{}  

\begin{abstract}
Streamflow forecasting is key to effectively managing water resources and preparing for the occurrence of natural calamities being exacerbated by climate change. Here we use the concept of fast and slow flow components to create a new mass-conserving Long Short-Term Memory (LSTM) neural network model. It uses hydrometeorological time series and catchment attributes to predict daily river discharges. Preliminary results evidence improvement in skills for different scores compared to the recent literature.
\end{abstract}

\section{Introduction}

Streamflow forecasting is essential in the planning and operation of water resources, such as supplying cities, electricity production, irrigation, and navigation. It also helps to prepare for natural disasters, such as floods and droughts. Having accurate methods to predict river discharge in advance would help to manage uncertainties that affect society and the economy, a challenge under climate change.
 
In this paper, we consider the problem of predicting next day streamflow given time series of past meteorological data (precipitation, soil moisture, and temperature) and physiographic attributes, such as land cover type, soil, topography, etc, of the catchment. We assume to have access to river discharge measurements for a certain period to train/calibrate the streamflow model. Once trained, the model should run on meteorological and physiographic data only, i.e, without access to new streamflow data to update the model. This simulates a scenario where a river gauge station ceased to operate, which is consistent with  
the globally observed decline in station coverage~\cite{essd-10-787-2018}.

Traditionally, streamflow forecasting is done using either empirical or conceptual models. The first seeks to establish mathematical relationships between streamflow and predictive variables, and are easier to conceive. Conceptual models seek to represent the knowledge of physical hydrological processes, and perhaps are easier to interpret. 

 
We follow the approach of empirical modeling, focusing on machine learning. Past solutions vary from simple auto-regressive models, such as multilinear regression (MLR) \cite{Tangborn1976}, auto-regressive moving average (ARMA) \cite{box2015time} and variations that attempt to model physical relations by using exogenous variables (ARMAX)~\cite{Haltiner2007}. While simpler models offer a very good solution for streamflow forecasting, often they require pre-processing steps before model application, such as time series seasonal decomposition.
 
%
%
Black-box approaches of the kind of artificial neural networks (ANN)~\cite{Govindaraju2000}, offers a straightforward alternative to model nonlinear hydrological relations. ANNs have been shown to outperform ARMA and MLR models~\cite{Hikmet2005}.  Under proper training, the prediction skill of many streamflow prediction approaches based on ANN improves as the flexibility (complexity) of the model increases, but at an increased computational cost to train. 
 
Recently, Long Short-Term Memory (LSTM), a type of artificial recurrent neural network architecture that can store and forget information over time has shown promising results in streamflow modeling~\cite{hess-23-5089-2019}.
 
Mass conservation is an important property exploited to customize LSTM formulations to ensure certain inputs are conserved and redistributed across storage locations in a system~\cite{hoedt2021mclstm}. In Hoedt’s et al. application in hydrology, the amount of water is conserved using an LSTM variant with new neural arithmetic units that preserve a constant sum of memory cells over time.
The  LSTM architecture use two types of inputs: one related to mass, such as  contributions to the streamflow, and auxiliary inputs, that variables control how the streamflow and these new contributions generate the next value of streamflow in a time series, acting as inputs to the gates that control mass fluxes.

In this paper we have three main objectives:
\begin{enumerate}
\item Propose a new LSTM architecture to estimate streamflow that implicitly modulate fast and slow flows components, respecting mass conservation.
\item Improve the efficiency of streamflow modeling by using a projection layer in our LSTM architecture, thus encoding catchment attributes in a smaller data space.
\item Test the skill of our LSTM architecture to predict streamflow in 32 stream gauges located in southern Brazil, comparing results to state-of-the-art models.
\end{enumerate}

\section{Related work}


\subsection{Mass-Conserving LSTM} \label{sec:MC-LSTM}
 \citet{hoedt2021mclstm} modified an LSTM to represent in a recurrent cell the dynamics of the water balance, by receiving water contributions $\mathbf{x}^{t}$ \eqref{eq:mass-balance},
\begin{align}
    \mathbf{m}_{tot}^{t} & = \mathbf{R}^{t}\cdot \mathbf{c}^{t-1}+ \mathbf{i}^{t} \cdot  \mathbf{x}^{t} \label{eq:mass-balance}\\
    \mathbf{c}^{t} & = (\mathbf{1}-\mathbf{o}^{t}) \odot \mathbf{m}_{tot}^{t} \label{eq:ceq}\\
    \mathbf{h}^{t} & = \mathbf{o}^{t} \odot \mathbf{m}_{tot}^{t} \label{eq:qeq}\\
    q^{t} &= \sum_{i=1}^{n-1} h_{i}^{t}
\end{align}
where:
\begin{align}
\mathbf{i}^{t} & = \text{softmax} \left( \mathbf{W}_{i} \cdot \mathbf{a}_{t} +\mathbf{U}_{i} \cdot \tilde{\mathbf{c}}_{t} + \mathbf{b}_{i}\right), \\
    \mathbf{o}^{t} & = \sigma\left( \mathbf{W}_{o} \cdot \mathbf{a}_{t} +\mathbf{U}_{o} \cdot \tilde{\mathbf{c}}_{t} + \mathbf{b}_{o}\right),\\
    \mathbf{R}^{t} & = \text{softmax} \left( \mathbf{W}_{r} \cdot \mathbf{a}_{t} +\mathbf{U}_{r} \cdot \tilde{\mathbf{c}}_{t} + \mathbf{B}_{r}\right),
\end{align}
and $\tilde{\mathbf{c}}_{t}$ is a normalization of $\mathbf{c}_{t}$ ($\nicefrac{\mathbf{c}_{t-1}}{\|\mathbf{c}_{t-1}\|}$). $\mathbf{W}$ and $\mathbf{U}$ are weight matrices, $\mathbf{b}$ and $\mathbf{B}_{r}$ are a bias vector that need to be learned during training. The $\mathbf{R}_{t}$ matrix regulates how much past water in the system is considered in the current water balance. This matrix depends on the remaining water of the system $\mathbf{c}^{t-1}$ and the forcing variables $\mathbf{a}_{t}$. Vector $\mathbf{i}^{t}$ controls the way that the auxiliary variables and the past water in the system influences the addition of new water.
Finally, vector $\mathbf{o}^{t}$ redistributes water in the system to the next streamflow water accumulators.  \cite{hoedt2021mclstm} found that this mass conserving LSTM works very well for extreme values of streamflow. 
 
\subsection{Fast and slow components of streamflow}

\citet{fastslow} proposed a simple model to predict 10-day streamflow totals in medium-sized watersheds using only average precipitation and soil moisture of the basins. Their model was proposed to estimate streamflow using only  data from the Soil Moisture Active Passive (SMAP) satellite mission. According to the authors, streamflow can be approximated by fast and slow flow components. The fast flow ($Q_{fast}$) portion of precipitation $p$ is proportional to the soil moisture $w$ of the top surface layers of the soil (i.e, $Q_{fast} 	\propto w \cdot p$), and the slow flow is $Q_{slow} \propto w$.
%
%
These relations can be combined to represent the dynamics of streamflow $Q$ as a quadratic function of $p$ and $w$:
\begin{equation}
\label{eq:fastflow}
	Q = Q_{fast} + Q_{slow} =  \alpha \cdot w \cdot p + \beta \cdot w + \gamma
\end{equation}
%
This simple model reasonably approximated streamflows in warm seasons. However, large biases were observed during periods of high rainfall~\cite{fastslow}.

\section{Fast-Slow Mass-Conserving LSTM Streamflow Model}
\label{sec:FS-LSTM}


The motivation to create our new LSTM model to forecast streamflow comes from the idea that precipitation ($p$) and soil moisture ($w$) contribute directly to the formation of new streamflow~\cite{fastslow}. Because of that, the name of the proposed architecture is FS-LSTM (Fast-Slow LSTM). We also take note of the mass-conserving LSTM model reviewed in Section~\ref{sec:MC-LSTM}, and we build on these prior works. 

\begin{figure}[tb]
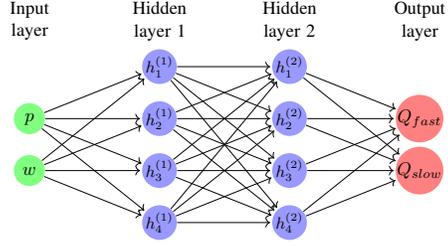

    \centering
    \resizebox{0.85\columnwidth}{!}{%
    \begin{neuralnetwork}[height = 4]
        \newcommand{\x}[2]{\ifnum1=#2 $p$ \else $w$ \fi}
        \newcommand{\y}[2]{\ifnum1=#2 $Q_{fast}$ \else $Q_{slow}$ \fi}
        \newcommand{\hfirst}[2]{\small $h^{(1)}_#2$}
        \newcommand{\hsecond}[2]{\small $h^{(2)}_#2$}
        \inputlayer[count=2, bias=false, title=Input\\layer, text=\x]
        \hiddenlayer[count=4, bias=false, title=Hidden\\layer 1, text=\hfirst] \linklayers
        \hiddenlayer[count=4, bias=false, title=Hidden\\layer 2, text=\hsecond] \linklayers
        \outputlayer[count=2, title=Output\\layer, text=\y] \linklayers
    \end{neuralnetwork}}
    \caption{Deep neural network $L(w,p)$ that will model the mass input in our FS-LSTM architecture.}
    \label{fig:deep}
\end{figure}

Equation~\eqref{eq:mass-balance2} models implicitly the fast and slow flow contributions, that depend on precipitation ($p^{t}$) and soil moisture ($w^{t}$) at day $t$:
\begin{equation}
    \mathbf{m}_{tot}^{t}  = \mathbf{R}^{t}\cdot \mathbf{c}^{t-1}+
    \mathbf{i}^{t}\cdot L(w^{t},p^{t}). \label{eq:mass-balance2}
\end{equation}
This means that instead of relying on the simple streamflow model considered in~\eqref{eq:fastflow}, we  account for possible nonlinearities in streamflow using a multilayer perceptron neural network $L(w^{t},p^{t})$, that takes as inputs precipitation and soil moisture and output the unknown fast and slow flow components.  Assuming that the relation between precipitation and soil moisture (modulating between fast and slow flow components) is close to the quadratic function proposed by~\cite{fastslow} shown in~\eqref{eq:fastflow}, we decided to use a relatively smaller number of layers and neurons in the network $L(w^{t},p^{t})$ shown in Figure~\ref{fig:deep}. To ensure positive values of $L(w^{t},p^{t})$ we use a quadratic function at the output layer. Also, using~\eqref{eq:ceq} and \eqref{eq:qeq} to determine $\mathbf{c}^{t}$ and $\mathbf{q}^{t}$, we have that:
\begin{align}
\mathbf{r}^{t} & = \mathbf{W}_{r} \cdot \mathbf{a}_{t}\\
\mathbf{i}^{t} & = \text{softmax} \left( \mathbf{W}_{i} \cdot \mathbf{r}^{t} +\mathbf{U}_{i} \cdot \hat{\mathbf{c}}_{t} + \mathbf{b}_{i}\right) \\
    \mathbf{o}^{t} & = \sigma\left( \mathbf{W}_{o} \cdot \mathbf{r}^{t} +\mathbf{U}_{o} \cdot \hat{\mathbf{c}}_{t} + \mathbf{b}_{o}\right)\\
    \mathbf{R}^{t} & = \text{softmax} \left( \mathbf{W}_{r} \cdot \mathbf{r}^{t} +\mathbf{U}_{r} \cdot \hat{\mathbf{c}}_{t} + \mathbf{B}_{r}\right)
\end{align}


Differently from the MC-LSTM model, we propose to use the sum of the past mass of the system $\hat{\mathbf{c}}_{t} = \sum_{i}^{N} c_{i,t}$. We found that this sum has a similar effect on the activation gates to $\tilde{\mathbf{c}}_{t}$ of the MC-LSTM, and $\hat{\mathbf{c}}_{t}$ has dimension one, thus reducing the number of weights. Furthermore, we use a projection layer $\mathbf{r}^{t}$ on the auxiliary inputs to reduce the dimensionality of the other activation layers. This projection acts like an encoder layer that compresses similar patterns into a smaller space compared to $\mathbf{a}^{t}$. This approach is similar to the approach used in~\cite{43895} to create a recurrent layer in the LSTM architecture that increases the number of units of the projection layers without increasing the number of parameters in all LSTM activations. 


\begin{figure}[tb]
\vskip 0.2in
  \centering
  \resizebox{\columnwidth}{!}{%
\begin{tikzpicture}[
    font=\sf \scriptsize,
    >=LaTeX,
    cell/.style={
        rectangle, 
        rounded corners=5mm, 
        draw,
        very thick,
        },
    operator/.style={
        circle,
        draw,
        inner sep=-0.5pt,
        minimum height =.4cm,
        },
    function/.style={
        ellipse,
        draw,
        inner sep=1pt
        },
    ct/.style={
        draw,
        line width = 0pt,
        minimum width=1cm,
        inner sep=1pt,
        },
    gt/.style={
        rectangle,
        draw,
        minimum width=4mm,
        minimum height=3mm,
        inner sep=1pt
        },
    mylabel/.style={
        font=\sffamily
        },
    ArrowC1/.style={
        rounded corners=.25cm,
        line width=0.75mm,
        },
    ArrowC2/.style={
        rounded corners=.5cm,
        line width=0.75mm,
        },
    ]

    \node [cell, minimum height =4.5cm, minimum width=7cm,fill=white!50!cyan] at (0,0){} ;

    
    \node [gt,fill=white!50!yellow] (ibox1) at (-2,-0.75) {softmax};

    \node [gt,fill=white!50!yellow] (ibox4) at (0.5,-0.75) {$\sigma$};
    \node [gt,fill=white!50!yellow] (ibox5) at (-1,-0.75) {softmax};

    \node [operator,fill=white!50!orange] (mux1) at (-2,1.5) {$\times$};
    \node [operator,fill=white!50!orange] (mux2) at (0,0) {$\times$};
    \node [operator,fill=white!50!orange] (add1) at (0,1.5) {+};
    \node [operator,fill=white!50!orange] (mux3) at (1.5,0) {$\times$};
    \node [operator,fill=white!50!orange] (min1) at (3,1.5) {--};

    \node[ct, label={[mylabel]In}] (c) at (-4.5,1.5) {\empt{c}{t-1}};
    \node (h) at (-3,-1.5){};
    \node[ct, label={[mylabel]left:Aux}] (a) at (-3,-3) {\empt{a}{t}};
    \node[ct, label={[mylabel]left:Input}] (x) at (0,-3) {\empt{(w,p)}{t}};
    \node[ct, label={[mylabel]Out}] (c2) at (4.5,1.5) {\empt{c}{t}};
    \node(h2) at (2.0,-1.5) {};
    \node[ct, label={[mylabel]left:Streamflow}] (x2) at (2.5,3) {\empt{q}{t}};

    \draw [ArrowC1] (c) -- (mux1) -- (add1) -- (min1) -- (c2);

    \draw [ArrowC1] (c) -- ( c-| h)|- (h);
    \draw [ArrowC1] (a) -- (a |- h)-| (ibox1);
    \draw [ArrowC1] (a) -- (a |- h)-| (ibox4);
    \draw [ArrowC1] (a) -- (a |- h)-| (ibox5);
    
    \draw (x |- h) ++(0,-0.1) coordinate (i2);
    \draw [-, ArrowC2] (x) -- (i2);
    \draw [-, ArrowC2] (i2)++(0,0.2) -- (mux2);
    \draw [->, ArrowC2] (ibox1) -- (mux1);
    \draw [->, ArrowC2] (mux2) -- (add1);
    
    \draw [->, ArrowC2] (ibox4) |- (mux3);
    \draw [->, ArrowC1] (add1 -| mux3)++(-0.5,0) -| (mux3);
    
     \draw [->, ArrowC2] (ibox5) |- (mux2);

    \draw [-, ArrowC2] (mux3) |- (h2);
    \draw (c2 -| x2) ++(0,-0.1) coordinate (i1);
    \draw [-, ArrowC2] (h2 -| x2)++(-0.5,0) -| (i1);
    \draw [-, ArrowC2] (i1)++(0,0.2) -- (x2);
    
    \draw [->,ArrowC2] (h2)++(-0.1,0) -| (min1);
 \node [gt,fill=white!50!yellow] (ibox2) at (-3,-2) {Lin};
  \node [gt, minimum width=0.8cm,fill=white!50!yellow] (ibox3) at (0,-2) {Deep};
\end{tikzpicture}
}
  \caption{Fast slow streamflow LSTM architecture.}
  \vskip -0.2in
\end{figure}
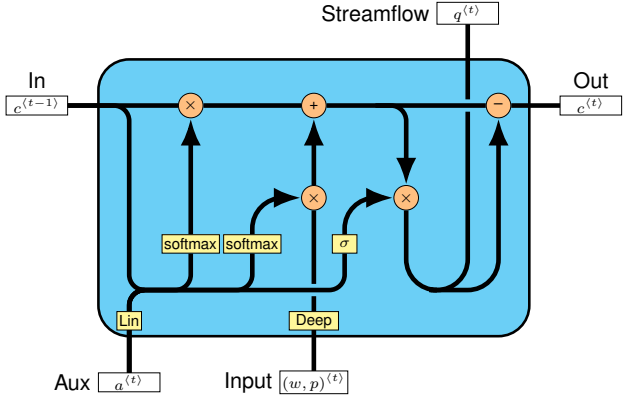

The MC-LSTM architecture has $2 \cdot n_{c} \cdot (n_{a}+n_{c})+n_{c}^{2} \cdot (n_{a}+n_{c})$ weights (without counting the biases), and our FS-LSTM has $2 \cdot n_{c} \cdot (n_{r}+n_{c})+n_{c}^{2} \cdot (n_{r}+n_{c})+n_{r} \cdot n_{a}$, where $n_c$ is the number of cells; $n_a$ is number of auxiliary inputs, and $n_r$ is the number of elements on the projected space. Our method can be trained using, for instance, fewer weights if we choose $n_r$ as:
\begin{equation} \label{eq:nrineq}
    n_{r} \leq n_{a} \cdot \frac{(n_{c}^{2}+2 \cdot n_{c})}{(n_{c}^{2}+2 \cdot n_{c}) +n_{a}}
\end{equation}
 
\section{Experiments}


\subsection{CAMELS-BR dataset and study area}
\label{sec:camelsbr}

\begin{figure}[tb]
\vskip 0.1in
    \centering
    \includegraphics[width=\columnwidth]{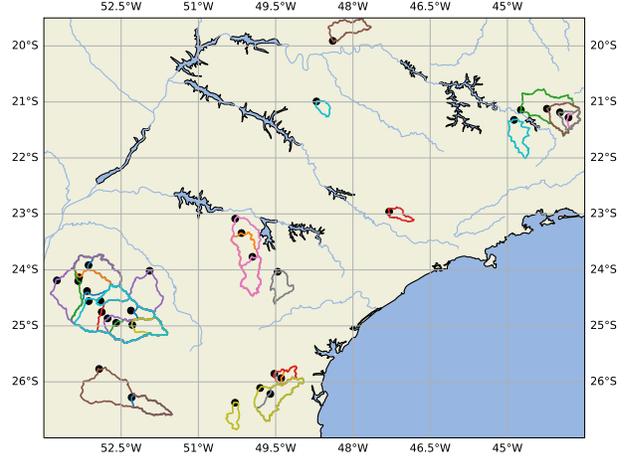}
    \caption{Selected stream gauge stations with their correspondent basins areas in southern Brazil.}
    \label{fig:camels_br_map}
\vskip -0.3in
\end{figure}
 
The data used in our research come from the freely available CAMELS-BR\footnote{Catchment Attributes and Meteorology for Large-sample Studies - Brazil  ~\cite{essd-12-2075-2020}.} dataset. We screen for the stream gauges located into coordinates the (54W,~19.5S) and (43.5W,~27S) of our study area shown in Figure~\ref{fig:camels_br_map}. We selected gauges having at least 10 years of quality-controlled daily time series of unregulated streamflow,  meteorological data, and catchment attributes.

We also included in the analysis soil moisture from gridded GLDAS~\cite{TheGlobalLandDataAssimilationSystem} version 2.0. This product provides the surface soil moisture daily average in kg/m$^2$ for the top 10-cm soil layer. We compute daily average soil moisture values using the polygon data available in the shapefiles (available in ~\cite{camelsbrlink}) that delineate the catchment of each stream gauge station.  

As $L(\cdot)$ input variable, we use the average of precipitation in the catchment, in mm/day, derived from CHIRPS~\cite{chirps}, that is distributed with CAMELS-BR. The auxiliary dynamic input variables were the minimum, mean, and maximum temperatures from NOAA~\cite{NOAA}. The static variables, used to characterize the catchment, are: elevation mean, slope mean, area, forest percentage, bedrock depth, water table depth, sand percentage, silt percentage, clay percentage, geological permeability, pressure mean, potential evapotranspiration mean, aridity, high precipitation frequency, high precipitation duration, low precipitation frequency and low precipitation duration.

\subsection{Evaluation criteria}
\label{sec:criteria}

We use six scores that attempt to capture different skills sought in streamflow predictions. The Nash–Sutcliffe model efficiency coefficient (\textbf{NSE})~\cite{NASH1970282} that helps to understand how the model improves over the mean. The Kling-Gupta Efficiency (\textbf{KGE})~\cite{GUPTA200980} which is a weighted sum of the three components that appear in the NSE formula: linear correlation, variability ratio, and bias ratio. The root mean squared error (\textbf{RMSE}). We also use three scores derived from the flow duration curve (FDC), a cumulative frequency curve that shows the percent of time specified discharges were equaled or exceeded during a given period (see examples in Figure~\ref{fig:minmax_nse}). The bias of the high-segment volume (\textbf{\%BiasFHV}) with exceedance probabilities 0.0-0.2, the mid-segment slope (\textbf{\%BiasFMS}) in 0.2-0.7, and the low-segment volume (\textbf{\%BiasFLV}) in 0.7-1.0 ~\citep{Yilmaz2008}.

\begin{table}[tb]
    \caption{Common setup for LSTM architectures.}
    \vskip 0.15in
    \begin{center}
\begin{small}
\begin{sc}
    \begin{tabular}{ccccc}
    \toprule
       \# cells  & \# epochs & batch & input & output \\
         \midrule
         64 & 30 & 256 & 365 & 1   \\
         \bottomrule
    \end{tabular}
    \label{tab:architecture}
    \end{sc}
\end{small}
\end{center}
\vskip -0.1in
\end{table}

\subsection{Experimental set-up}

%

All stations shared in common daily time series from 1 October 1994 to 30 September 2008.  We trained on streamflow observations from 1 October 1999 to 30 September 2008  and tested on observations from 1 October 1994 to 30 September 1999. The remaining initial observations (1 October 1990 to 30 September 1994) were set aside for validation during the training to avoid over-fitting problem.

We compared our FS-LSTM model to three alternative methods: a vanilla LSTM \cite{Kratzert2018}; EA-LSTM~\cite{hess-23-5089-2019}, in both cases the meteorological forcing data and additional catchment attributes controls the state space that are used; and  MC-LSTM~\cite{hoedt2021mclstm} considered state-of-the-art. The common setup configuration for all the lstm architectures is shown in Table~\ref{tab:architecture}. In this table, the input size includes the number of previous time steps of mass, and auxiliary catchment variables that are needed to forecast the streamflow output. For the ANN architecture shown in Figure~\ref{fig:deep}, we used 2 layers and 10 neurons per layer. Also for we used $n_r = 10$ that respect~\eqref{eq:nrineq}. In our experiments with MC-LSTM, we set  soil moisture as an input mass variable, not originally considered in~\cite{hoedt2021mclstm}. Each model trains with all gauges pooled together in 30 epochs. 




\subsection{Results}

\begin{figure}[tb]
    \centering
    \resizebox{\columnwidth}{!}{\input{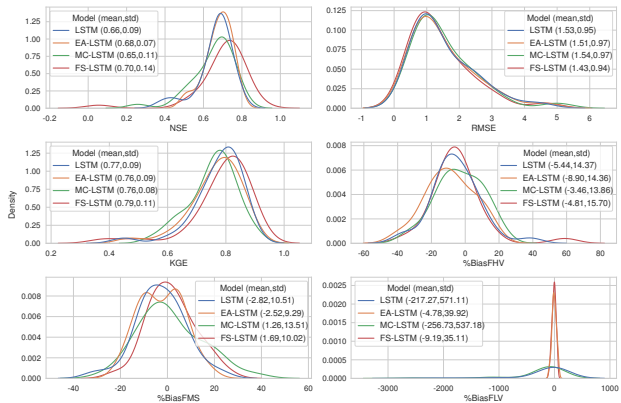} }
    \caption{Metrics comparing different streamflow models in 32 CAMELS-BR stream gauges.}
    \label{fig:camelsbr_benchs}
\vskip -0.2in
\end{figure}

 The charts in Figure~\ref{fig:camelsbr_benchs} illustrate the kernel density estimates for the six scores evaluated in the 32 basins during the testing period. The proposed FS-LSTM outperforms the alternative algorithms tested in terms of NSE, with an average of 0.7 compared to 0.68 of EA-LSTM, ranked second, and 0.66 of the vanilla LSTM, with relatively higher standard deviations. A relative improvement (RI) of 2.9\%. FS-LSTM achieved the lowest average RMSE error of 1.43 mm/day, compared to 1.51 mm/day (5.3\% RI) of EA-LSTM, all having similar density curves. In terms of KGE, the advantage of FS-LSTM also becomes clear, with a density curve shifted towards higher values, with an average of 0.79 compared to the 0.77 (2.6\% RI) of the vanilla LSTM.

Results are more mixed for the scores based on the FDC. MC-LSTM outperforms the other methods in terms of average \%BiasFHV, but we note that all models perform well since most of their densities values have bias within the $\pm$25$\%$ range recommend by~\cite{Moriasi2007}. With \%BiasFMS, arguably the proposed FS-LSTM has the second overall performance because the kernel density curve looks more symmetrical around zero and only a small portion of the tales are outside the target $\pm$25$\%$ bias bound. For the \%BiasFLV, our experiments suggest that FS-LSTM and EA-LSTM perform very well compared to MC-LSTM and LSTM, attaining the desired $\pm$25$\%$ bound.

%
%

\begin{figure}[tb]
    \centering
    \resizebox{\columnwidth}{!}{\input{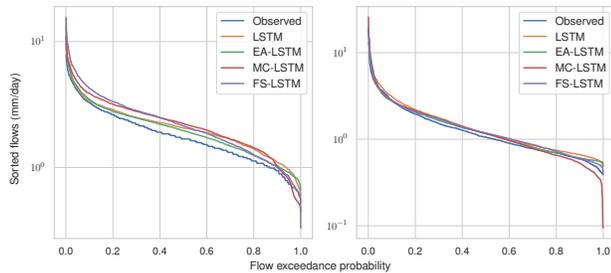} }
    \caption{Flow Duration Curve for basins with the worst (left) and best (right) NSE.}
    \label{fig:minmax_nse}
\end{figure}

Figure~\ref{fig:minmax_nse} presents two examples of FDC for the gauges with the worst and best NSE skill over all models, from left to right, respectively. These curves show the percentage of time a certain river discharge was equaled or exceeded during the testing period. Even in the worst case, FS-LSTM provided competitive results in the flow exceedance probability range $\approx$~0.2-0.9.  
 
\section{Conclusions}

The approach proposed herein exploits a novel strategy that uses a multilayer perceptron artificial neural network to implicitly model the fast and slow streamlfow components in a modified mass-conserving LSTM. We demonstrate that the proposed FS-LSTM achieves high prediction skill for gauges located in southern Brazil. Improvements in the low streamflow volumes remain a challenge, as well as the investigation of strategies to transfer these models to other geographies with less retraining effort.  
  
\bibliographystyle{icml2021}

\end{document}